\documentclass{article}

\usepackage{PRIMEarxiv}

\usepackage[utf8]{inputenc} 
\usepackage[T1]{fontenc}    
\usepackage{hyperref}       
\usepackage{url}            
\usepackage{booktabs}       
\usepackage{amsfonts}       
\usepackage{nicefrac}       
\usepackage{microtype}      
\usepackage{lipsum}
\usepackage{fancyhdr}       
\usepackage{graphicx}       
\graphicspath{{media/}}     

\usepackage{times} 
\usepackage{amsmath} 
\DeclareMathOperator*{\argmax}{arg\,max}

\usepackage{amssymb}  
\usepackage{subfig}
\usepackage{booktabs}
\usepackage{multirow}
\usepackage{enumitem}
\title{Position-Agnostic Autonomous Navigation in Vineyards with Deep Reinforcement Learning
}

\author{Mauro Martini$^{1,2}$, Simone Cerrato$^{1,2}$, Francesco Salvetti$^{1,2,3}$, Simone Angarano$^{1,2}$, and Marcello Chiaberge$^{1,2}$
\thanks{$^{1}$Department of Electronics and Telecommunications, Politecnico di Torino, Turin, Italy
{\tt\footnotesize (name.surname@polito.it)}}%
\thanks{$^{2}$PIC4SeR Interdepartmental Center for Service Robotics, Politecnico di Torino, Turin, Italy, https://pic4ser.polito.it}%
\thanks{$^{3}$SmartData Interdepartmental Center for Big Data and Data Science, Politecnico di Torino, Turin, Italy, https://smartdata.polito.it}%
}

\begin{document}
\maketitle

\begin{abstract}
Precision agriculture is rapidly attracting research to efficiently introduce automation and robotics solutions to support agricultural activities. Robotic navigation in vineyards and orchards offers competitive advantages in autonomously monitoring and easily accessing crops for harvesting, spraying and performing time-consuming necessary tasks. Nowadays, autonomous navigation algorithms exploit expensive sensors which also require heavy computational cost for data processing. Nonetheless, vineyard rows represent a challenging outdoor scenario where GPS and Visual Odometry techniques often struggle to provide reliable positioning information. In this work, we combine Edge AI with Deep Reinforcement Learning to propose a cutting-edge lightweight solution to tackle the problem of autonomous vineyard navigation without exploiting precise localization data and overcoming task-tailored algorithms with a flexible learning-based approach. We train an end-to-end sensorimotor agent which directly maps noisy depth images and position-agnostic robot state information to velocity commands and guides the robot to the end of a row, continuously adjusting its heading for a collision-free central trajectory. Our extensive experimentation in realistic simulated vineyards demonstrates the effectiveness of our solution and the generalization capabilities of our agent.
\end{abstract}


\section{Introduction}
In recent years, the development of Agriculture 3.0 and 4.0 paradigms rapidly attracted research attention with the aim of satisfying four essential requirements: increasing productivity, allocating resources reasonably, adapting to climate change, and avoiding food waste \cite{zhai2020decision}. One fundamental step for introducing an efficient and reliable automation in the agriculture processes is the development of a robotic autonomous navigation pipeline. This is the first requirement to successfully design a platform that takes care of several tasks such as harvesting \cite{harvesting}, spraying \cite{spraygrape, deshmukh2020design}, vegetative assessment and yield estimation \cite{zhang2020assessment, feng2020yield}, and many others \cite{seeding, irrigation}. In this work, we focus on the autonomous navigation of an Unmanned Ground Vehicle (UGV) in row crops and, in particular, we develop and test our methodology in a realistic simulated vineyard.

\begin{figure}[ht]
\centering
\includegraphics[width=0.8\columnwidth]{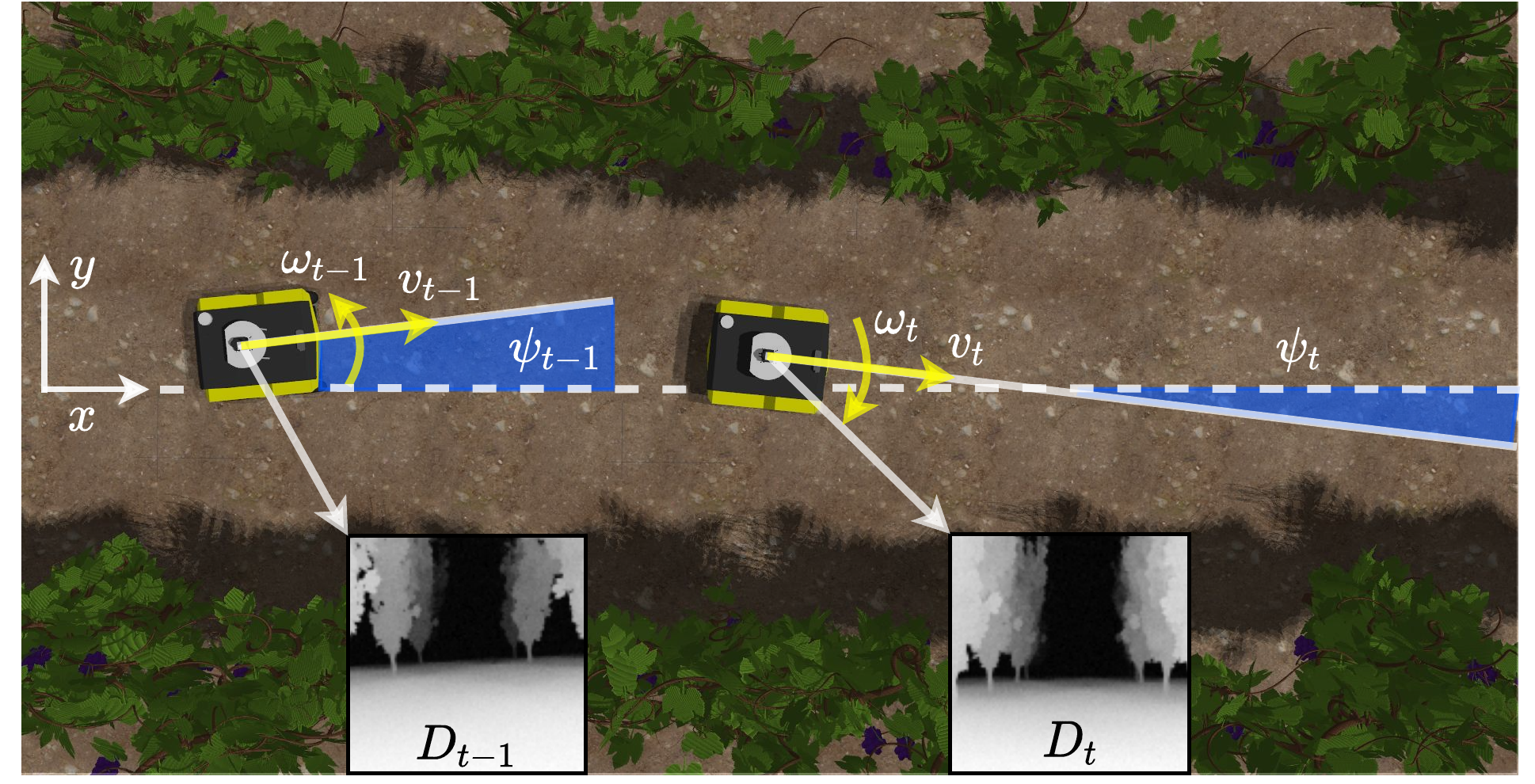}
\caption{
At each time instant $t$, the proposed agent receives as inputs a raw noisy depth image $D_{t}$, the previous velocity commands $[v_{t-1},\omega_{t-1}]$ and the yaw $\psi_{t}$, to generate the new commands $[v_t,\omega_t]$. Since no explicit localization is required, the agent performs a positioning-independent navigation in the vineyard row.}
\label{fig:observations}
\end{figure}

\subsection{Autonomous Navigation in Precision Agriculture}

Nowadays, autonomous navigation in agricultural contexts is tackled with the usage of expensive high-precision GNSS sensors, such as GPS receivers with RTK corrections \cite{thuilot2001automatic, gps_only}, usually in combination with laser sensors \cite{lidar_gps2,callegati2018autonomous}. However, the presence of thick canopies and lush vegetation decreases the reliability of GNSS sensors \cite{marden2014gps, kabir2016performance}, especially during spring and summer. This condition strengthens the need for alternatives to reduce the cost of the system without affecting its robustness. Visual odometry \cite{aguiar2019monocular, zaman2019cost, nevliudov2021development} and computer vision methods \cite{ma2021autonomous} have been proposed as different valid options to tackle navigation inside fields. However, these methods generalize poorly in the case of long outdoor paths with repetitive visual patterns.

Deep Learning (DL) methods are more and more used in the context of precision agriculture \cite{KAMILARIS201870, ren2020survey} to tackle many different tasks such as object detection \cite{mazzia2020real, bazame2021detection}, fruit counting \cite{rahnemoonfar2017deep}, land crop classification \cite{martini2021domain, xu2021towards}. Recently, DL solutions have been proposed to solve the autonomous navigation problem, overcoming the limits of localization in row crops scenarios with methods of orientation classification \cite{aghi2020local} and plant segmentation \cite{aghi2021deep, cerrato2021deep}. 

\subsection{Robot Navigation with Deep Reinforcement Learning}
Classical navigation algorithms execute perception, planning and control as separate concatenated stages, increasing the overall risk of error in each sub-module. Differently, policy learning methods can directly map raw input data to actions, drastically simplifying the whole navigation algorithmic pipeline.
Model-free Deep Reinforcement Learning (DRL) optimizes a parametric policy without accessing the dynamic model of the environment, allowing to train an agent to navigate in unseen environments. DRL has recently emerged as a powerful approach for autonomous vehicles \cite{aradi2020survey} and mobile robot navigation \cite{zhu2021deep}. Whilst the majority of DRL works in literature focuses on indoor local planning \cite{mirowski2016learning} or obstacle avoidance with 2D lidars \cite{leiva2020robust} or cameras \cite{choi2019deep}, recent studies have chosen DRL agents to tackle more challenging tasks such as socially-aware motion planning \cite{chen2017socially} and multi-robot systems \cite{fan2020distributed}.

However, vineyard navigation presents considerable critical aspects with respect to point-to-point navigation in open-space scenarios. Firstly, the constrained geometry of the vineyards may cause episode early stopping for collision, limiting the amount of states visited before and during the training of the policy. Similarly to the approach adopted for high-speed drone racing \cite{song2021autonomous}, we tackled the problem by continuously changing the starting point of the robot in the training stages. Moreover, assuming no information about the robot position is available, the input state does not include key information on the distance or heading with respect to the goal.

\subsection{Contributions}
\begin{figure}[ht]
    \centering
    \includegraphics[width=0.7\columnwidth]{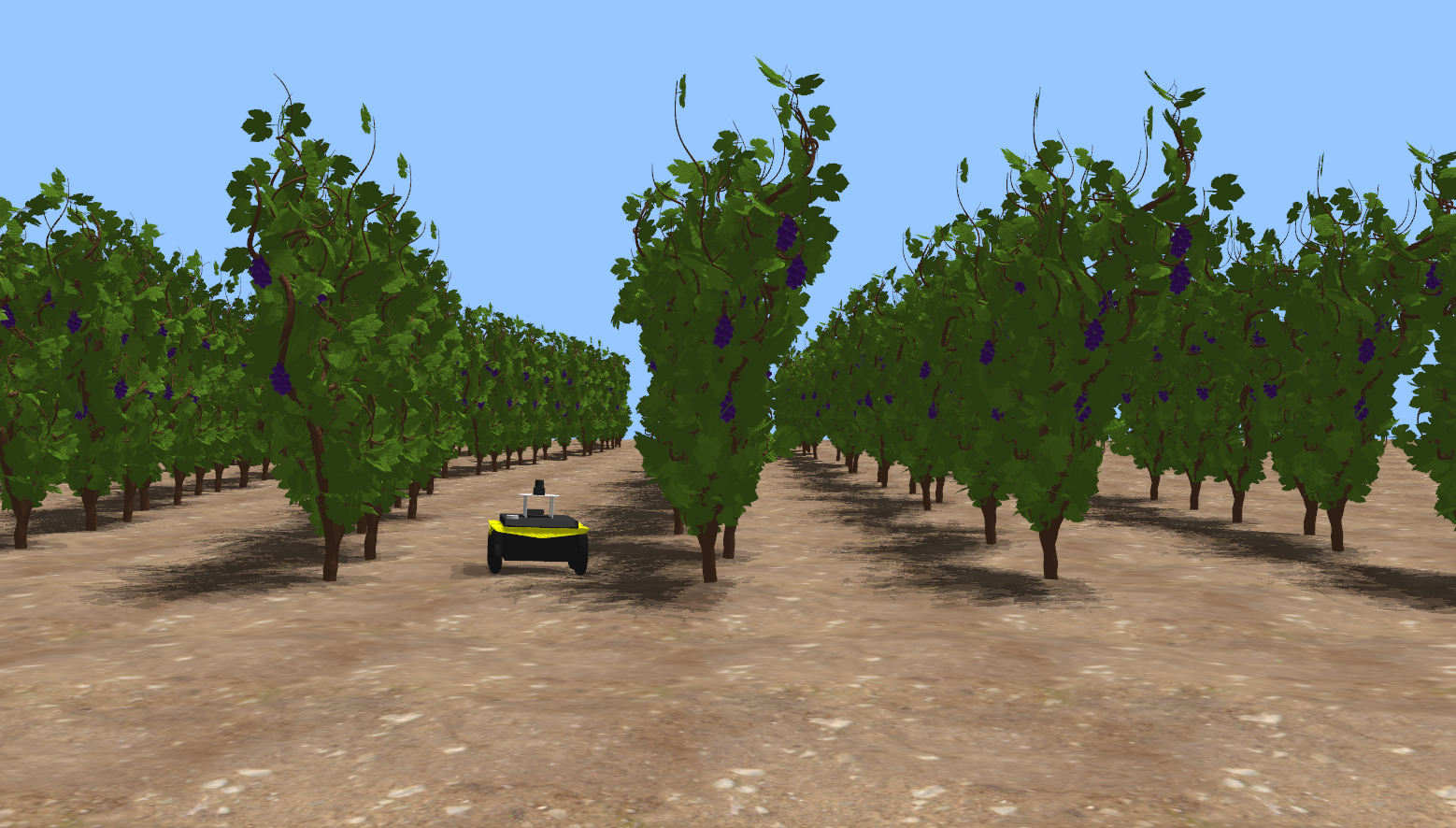}
    \caption{The robot is placed in a vineyard environment with realistic terrain profile. In testing rows, curves and plant gaps are introduced to validate the agent in conditions comparable to real-world challenging scenarios.}
    \label{fig:test_perspective}
\end{figure}

In this paper, we propose a novel method to tackle navigation inside row crops, based on a sensorimotor DRL agent that directly maps a raw depth image combined with a basic robot state composed of yaw and previous velocity commands to new actions, as shown in Fig. \ref{fig:observations}. This approach does not need any precise localization nor any information on the actual distance from the goal and can therefore be used when GPS and visual-based positioning are unreliable. In particular, we focus on the intra-row segments, since they are the most affected by GNSS signal deterioration, especially with lush vegetation and thick canopies. However, our approach can be easily combined with a waypoint-based global planner \cite{mazzia2021deepway} to achieve a complete field coverage, tackling inter-row paths with classical approaches, such as in \cite{cerrato2021deep}. The proposed DRL agent is trained in simulation on custom environments specifically designed with the aim of reproducing the geometry of real vineyards. An example of such simulated scenario is shown in Fig. \ref{fig:test_perspective}.

To summarize, the main advantages of the proposed method are the following:
\begin{itemize}
\item the model is position-agnostic and can reach the end of the row without global localization;
\item the vision and control systems are coupled together and the policy learns a direct mapping between sensor-state vectors and motion during training (sensorimotor agent);
\item the controller is environment-independent, since there is no need for adaptations to deal with unseen scenarios (e.g. curved rows);
\item the system is computationally efficient and it is optimized to run in real-time directly on-board of the robotic platform.
\end{itemize}

\section{METHODOLOGY}
\subsection{Task Formulation}
We model the navigation problem according to a reinforcement learning framework. Therefore, the problem is formulated as a Markov Decision Process (MDP) described by the tuple $(\mathcal{S},\mathcal{A}, \mathcal{P}, R, \gamma)$ \cite{sutton2018reinforcement}. An agent starts its interaction with the environment in an initial state $s_0$, drawn from a pre-fixed distribution $p(s_0)$ and then cyclically select an action $\mathbf{a_t} \in \mathcal{A}$ from a generic state $\mathbf{s_t} \in \mathcal{S}$ to move into a new state $\mathbf{s_{t+1}}$ with the transition probability $\mathcal{P(\mathbf{s_{t+1}}|\mathbf{s_t},\mathbf{a_t})}$, receiving a reward $r_t = R(\mathbf{s_t},\mathbf{a_t})$.

The aim of a reinforcement learning process is to optimize a parametric policy $\pi_\theta$ which defines the agent behaviour once trained. In the context of autonomous navigation, we model the MDP with an episodic structure with maximum time steps $T$, hence the agent is trained to maximize the cumulative expected reward $\mathbb{E}_{\tau\sim\pi} \sum_{t=0}^{T} \gamma^t r_t$ over each episode, where $\gamma \in [0,1)$ is the discount factor. More in detail, we use a stochastic agent policy in an entropy-regularized reinforcement learning setting, in which the optimal policy $\pi^*_\theta$ with parameters $\mathbf{\theta}$ is obtained maximizing a modified discounted term:

\begin{equation}
    \pi^*_\theta = \argmax_{\pi} \mathbb{E}_{\tau\sim\pi} \displaystyle \sum_{t=0}^{T} \gamma^t [r_t + \alpha \mathcal{H}(\pi(\cdot|s_t))]
\end{equation}

Where $\mathcal{H}(\pi(\cdot|s_t))$ is the entropy term which increases robustness to noise through exploration, and $\alpha$ is the temperature parameter which regulates the trade-off between reward optimization and policy stochasticity. 

\subsection{Reward} 
\begin{figure}[t]
\centering
\includegraphics[width=0.8\columnwidth]{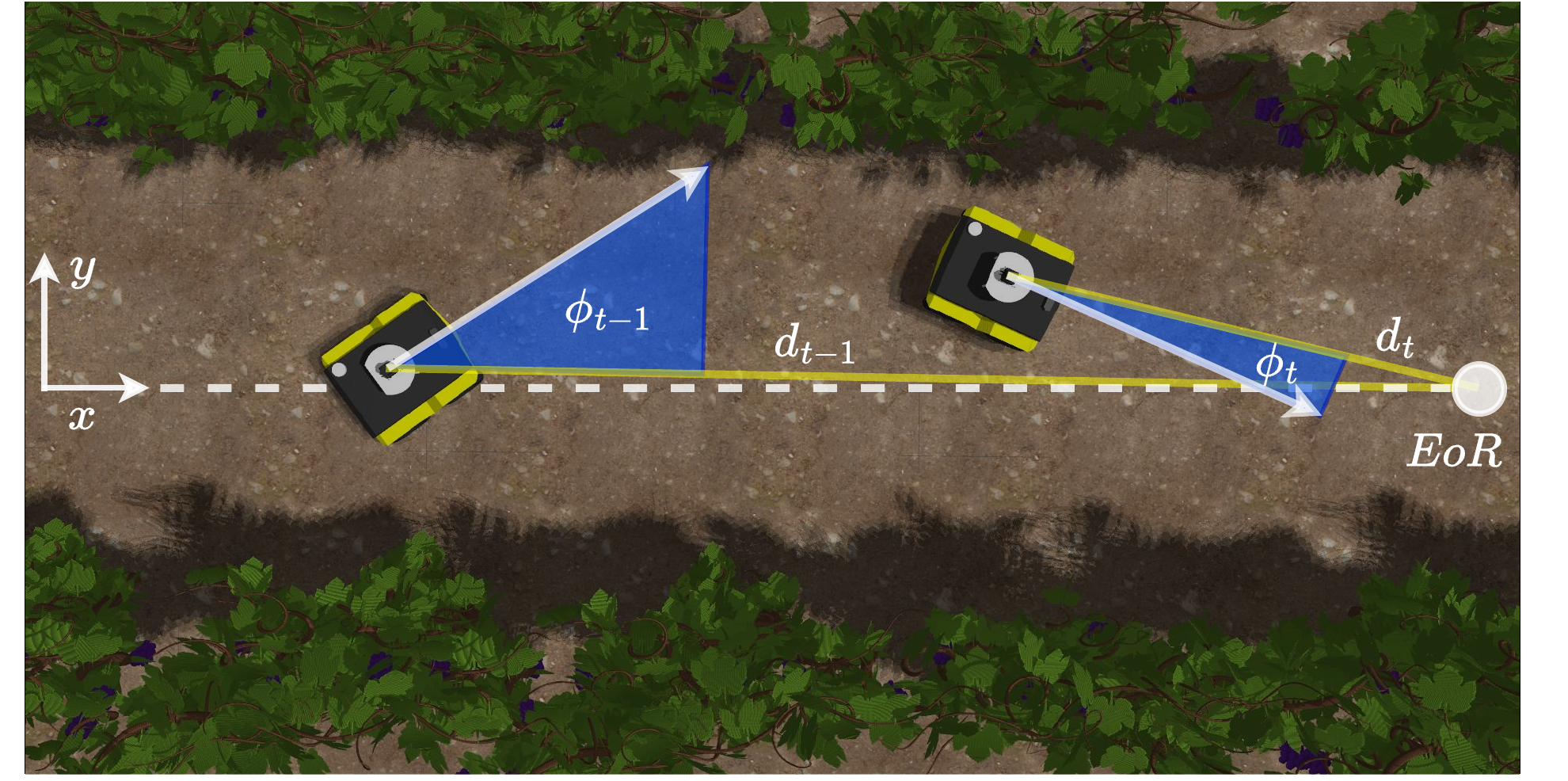}
\caption{The reward at each time step $t$ is computed as a function of the distances from the end of the row (EoR) $d_t$ and $d_{t-1}$, and the angle $\phi_t$ between the robot orientation and the shortest path to EoR. This information is available while training the agent although it does not constitute its input.}
\label{fig:reward}
\end{figure}

The potential outcome of the vineyard navigation task can be easily summarized in a binary output: the robot successfully arrives at the end of the row or it does not. However, this sparse reward feedback can be assigned to the agent only in the case of a completed successful episode, which is an improbable event all over the initial learning phases.
According to this, reward shaping is the typical process which leads researchers to analytically specify the desired behaviour to the agent thanks to a dense reward signal assigned at each time step. Moreover, this approach allows to precisely express secondary desired behaviours. In this application scenario, we identify three key features of an ideal optimal policy: a complete collision-free travel in the vineyard row, a centered trajectory and a proper orientation. Nonetheless, the reward can be computed during the training in simulation exploiting positioning data that is not needed at test time by the agent. 
To this end, we first define a reward contribution $r_h$ to keep the robot oriented towards the end of the row:
\begin{equation}
r_h = \left( 1 - 2 \sqrt{\left| \frac{\phi}{\pi} \right|} \right)
\end{equation}
where $\phi$ is the heading angle of the robot, namely the angle between its linear velocity and the end of the row. We consider $r_h$ as a fundamental feedback to let the agent understand how to counteract the sudden angular deviations imposed by the irregular terrain, which is realistically generated in our simulated world. Then, in order to obtain a central trajectory, we consider the possibility of directly scoring the distance of the robot from the center of the row. However, this approach requires to compare the robot pose with the mean line of each specific row at each step, and nonetheless it results in a slow and inefficient policy when combined with the heading reward $r_h$. For this reason, we prefer a distance-based reward to strongly encourage the agent to reach the end of the row:
\begin{equation}
    r_{d} = d_{t-1} - d_t 
\end{equation}
where $d_{t-1}$ and $d_t$ are euclidean distances between the robot and the end of the row (EoR) at successive time steps, as shown in Fig. \ref{fig:reward}. Robot pose information is uniquely used for reward computing while training, and is not included as agent input, as better specified in the following subsection \ref{subsec:policy}.
We finally include a sparse reward contribution for end-of-episode states, assigning $r_s = 1000$ for the successful completion of the task, $r_c = -500$ if collision occurs, and $r_\psi = -500$ if the robot overcomes a $\pm85^{\circ}$ yaw limit. Stopping the episode when the robot exits the vineyard row or reverses its motion direction is fundamental to keep collecting meaningful sample transitions for the task.

The final reward signal results to be as follows:
\begin{equation}
r = a \cdot r_h + b \cdot r_{d} + \begin{cases} 
                 r_s \; \;  \textit{if Success} \\
                 r_c \; \; \textit{if Collision} \\
                 r_\psi \; \; \textit{if Reverse} \\
            \end{cases}
\end{equation}
Where $a=0.6$ and $b=35$ are numerical coefficients to efficiently integrate the diverse reward contributions in the final signal.

\subsection{Policy Network}
\label{subsec:policy}
We define the parametrized agent policy with a deep neural network. We train the agent with the Soft Actor-Critic (SAC) algorithm presented in  \cite{haarnoja2018soft}, which allows for a continuous action space. In particular, we instantiate a stochastic Gaussian policy for the actor and two Q-networks for the critics.

\begin{figure*}[ht]
\centering
\includegraphics[width=0.8\textwidth]{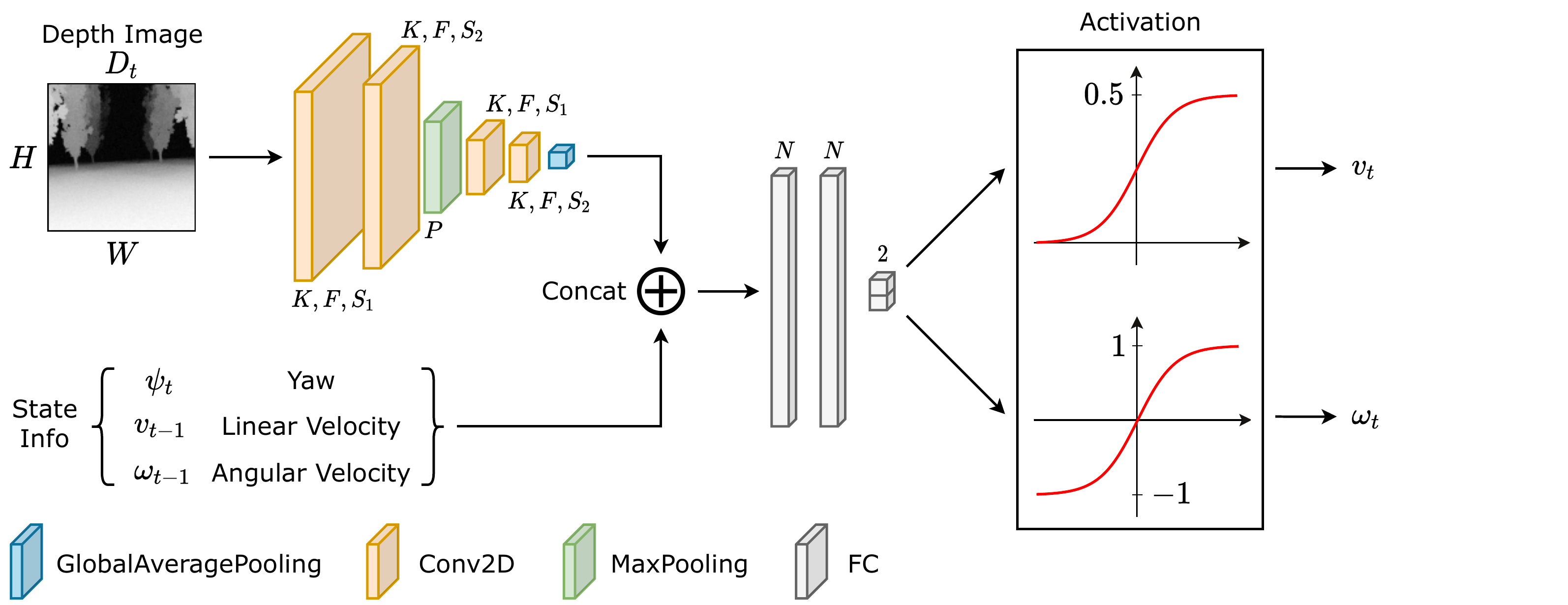}
\caption{Architecture of the actor policy network. A convolutional backbone extracts features from the depth image $D_{t}$. The features are then concatenated to the input vector $[v_{t-1},\omega_{t-1},\psi_{t}]$ and two fully connected layers output the action vector $[v_t,\omega_t]$ with specific activation functions.}
\label{fig:architecture}
\end{figure*}

\begin{description}[leftmargin=0pt]
\item[Input features] We select the input features of the policy network only considering the odometric and perception data available during the vineyard navigation task. To this end, we identify a set of input features which enables a localization-independent navigation, and an affordable perception system such as a simple depth camera. Several iterations lead us to the choice of three key elements as agent inputs.

1) The previous set of velocity commands $[v_{t-1}, \omega_{t-1}]$ to provide information about its current motion and temporal continuity to the agent and avoid strongly oscillating or disentangled commands.

2) The yaw of the robot $\psi_t$ measured at the current time instant $t$, which is always available thanks to an IMU sensor on the robot. The simple yaw angle does not represent the required optimal heading angle $\phi_t$ for the task. Indeed, such angle depends on the robot position, which is not available at test time in GPS-denied conditions. However, the yaw $\psi_t$ provides awareness about the actual orientation of the robot and helps in generating smooth collision-free velocity commands.

3) A raw noisy depth image of size $(112\times112)$. Each pixel of the image contains a distance in the range of $[0.0, 5.0] m$ and it is the only perception data of the environment the agent exploits to guide the robot through the vineyard. Processing the image, the agent acquires the necessary knowledge on obstacles and it can also visually infer its current orientation with respect to the end of the visible row. A reduced size of the image allows to obtain a compact latent representation of only 32 features. The choice of depth images is also motivated by the aim of reducing the simulation-to-reality gap. Indeed, depth images are only marginally affected by visual features. On the other hand, real depth images may present peculiar noisy behaviours, so we add to each raw image two different noises: a first uniform random noise with values in $[-0.5, 0.5] m$, and a second random noise proportional to the depth values in the image, also in the range $[-0.5, 0.5] m$. The second noise aims at disturbing more heavily the higher-distance points. Overall, each pixel presents at most a perturbation of $\pm 1 m$.

\item[Output actions]
The policy network predicts an action $a_t = [v_t, \omega_t]$ at each time step, which directly represents the required linear and angular velocity command to control the robot. This choice is mainly dictated by the differential-drive steering model of the robotic platforms we use, and it also allows to easily integrate the system with ROS standard command messages.

\item[Network architecture]
The architecture of the policy network presents a double input structure, represented in Fig. \ref{fig:architecture}. A convolutional feature extractor is designed to efficiently map the depth image in a compact latent representation, inspired by the work proposed for visual SAC in \cite{yarats2021improving}. However, we avoid the adoption of an encoder-decoder structure and an additional reconstruction loss. Our solution allows the agent to easily learn how to extract relevant feature for the task autonomously. 
Firstly, the feature extractor takes the depth image $D_{t}$ ($H\times W$) as input. It consists of two convolutional layers (kernel size $K=3$, $F=32$ filters, strides $S_1=2$ and $S_2=1$ respectively) with ReLU activations followed by a max pooling layer (pool size $P=2$), two more convolutional layers (with the same structure as the previous ones), and a global average pooling layer. The features are then concatenated to the position-agnostic robot state input vector, which includes the measured yaw $\psi_{t}$ and the previous action command (linear velocity $v_{t-1}$ and angular velocity $\omega_{t-1}$). 
The resulting vector is processed by two fully connected layers with $N=256$ neurons and ReLu activation. Finally, an additional output layer predicts the action vector $[v_t, \omega_t]$, using \textit{tanh} as activation function. The linear velocity $v_t$ is further squashed to $(0,0.5)$ to match the velocity profile of the robot.
As we use a stochastic Gaussian policy, the network actually predicts the mean and the standard deviation of each action distribution, which are used to sample a value from the derived distribution while training. Instead, the mean value is directly used as output at test time. The critic network structure is identical to the actor one, except it takes also the predicted action vector in input and outputs an estimate of the $Q$ value and it is trained according to the SAC algorithm. 

\begin{table*}[ht]
\centering
\caption{Performance of the agent in the test environment (forward F and reverse R). We include both MAE and RMSE metrics (lower is better), the actions $v$ and $\omega$ (mean and standard deviation) and the success rate (higher is better). The last row reports the overall mean results and metrics.}
\label{tab:test_results}
\resizebox{0.8\textwidth}{!}{%
\begin{tabular}{@{}ccccccccc@{}}
\toprule
Test Row & Row Shape & MAE {[}m{]} & RMSE {[}m{]} & $v$ {[}m/s{]} & $\omega$ {[}rad/s{]} & Success\\ \midrule
1F &Straight & 0.076 & 0.089 & 0.493 ± 0.020 & -0.027 ± 0.538 & 5/5\\
1R &Straight & 0.131 & 0.141 & 0.478 ± 0.053 & -0.030 ± 0.623 & 5/5\\\midrule
2F &Straight & 0.068 & 0.076 & 0.488 ± 0.037 & -0.028 ± 0.642 & 5/5\\
2R &Straight & 0.087 & 0.105 & 0.489 ± 0.037 & -0.001 ± 0.610 & 5/5\\\midrule
3F &Hybrid & 0.120 & 0.154 & 0.497 ± 0.003 & -0.012 ± 0.353 & 5/5\\
3R &Hybrid & 0.100 & 0.130 & 0.494 ± 0.017 & -0.036 ± 0.610 & 2/5\\\midrule
4F &Curved & 0.164 & 0.191 & 0.468 ± 0.079 & -0.008 ± 0.687 & 5/5\\
4R &Curved & 0.081 & 0.097 & 0.484 ± 0.059 & -0.014 ± 0.493 & 5/5\\\midrule
5F &Curved & 0.112 & 0.145 & 0.486 ± 0.057 & -0.005 ± 0.571 & 3/5\\
5R &Curved & 0.201 & 0.233 & 0.445 ± 0.122 & -0.042 ± 0.656 & 5/5\\\midrule
\textbf{Overall} & - & \textbf{0.114} & \textbf{0.136} & \textbf{0.482 ± 0.048}	& \textbf{-0.020 ± 0.578} & \textbf{45/50}\\ \bottomrule
\end{tabular}%
}
\end{table*}

\item[Random Initialization Strategy]
A critical condition of the vineyard environment is its constrained geometry. As a consequence, the agent usually collides in few steps in the early stage of the DRL training, exploring a drastically reduced number of states of the environment. This behaviour negatively affects the generalization properties of the agent and more generally the probability of a stable convergence of the training algorithm. For this reason, we identify a set of counteractions to effectively train the policy:
1) The vineyard training stage comprises different vineyard rows where the robot travels during the same training simulation to encourage better generalization derived from a higher number of visited states.
2) The robot is initialized in a new random pose in the vineyard every $10$ episodes, varying its $[x_0,y_0]$ position coordinates and its initial yaw $\psi_0$, enabling also the agent to travel the rows in both direction. Consequently, the agent is able to visit the final sections of rows from the beginning of the simulation, significantly speeding up the convergence of the training with better generalization results.
3) An initial exploration phase is combined with an additional $\epsilon$-greedy policy which samples random values in the action spaces with a probability that is exponentially reduced with the increase of episodes to maintain a proper level of exploration during the whole training.
\end{description}


\section{EXPERIMENTAL SETTING}
\label{sec:experiments}
In this section we present how the DRL agent is trained, detailing all the hyperparameters involved in the training procedure. Moreover, we present the custom simulation environments, specifically designed to validate the methodology.

\subsection{Training Setting}
The agent is customized starting from the implementation in the TF2RL library \footnote{https://github.com/keiohta/tf2rl}. Actor and critic networks present $104,100$ and $103,841$ parameters respectively and are trained with Adam optimizer and a learning rate of $2\cdot10^{-4}$. The $\epsilon$-greedy exploration policy is defined by the starting value $\epsilon_0 = 1.0$, the decay $\gamma_\epsilon=0.992$ and a minimum value for random action sampling of $\epsilon_{min} = 0.05$. The agent is trained for $1500$ episodes composed of $T=700$ maximum steps, changing the robot pose in different vineyard training rows every $10$ episodes, to guarantee a good level of exploration and resulting generalization.

Both training and testing are performed exploiting the open-source simulator Gazebo\footnote{http://gazebosim.org/}. It is compatible with ROS2 and has very good performances in terms of dynamic simulation capabilities as well as sensors simulation. Indeed, all the code is developed in compliance with ROS2 standards and is tested on Ubuntu 20.04 LTS using ROS2 Foxy.

\subsection{Simulation Environments}
The simulation environments are composed of three main models: the URDF model of the rover, our custom model of vine plant and the terrain model. The first contains all the useful information about the mechanical structure as well as links and joints among the different parts of the hardware platform. During the training phase, we adopt the model of the Jackal UGV\footnote{https://clearpathrobotics.com/jackal-small-unmanned-ground-vehicle/}, provided by Clearpath Robotics. The plant models are randomized in terms of orientation and deviation from their ideal positions in the rows. The model of the terrain is obtained using the heightmap option of the SDF format to obtain a bumpy and uneven simulated terrain. This allows to reproduce the sudden steerings and the jerky camera movements typical of a realistic interaction.

\begin{figure}[t]
    \centering
    \subfloat[]
    {
        \includegraphics[width=0.7\columnwidth]{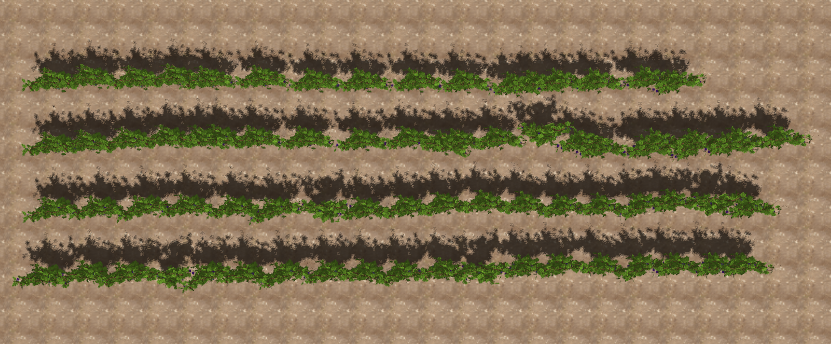}
        \label{fig:train_env}
    }\\
    \subfloat[]
    {
        \includegraphics[width=0.7\columnwidth]{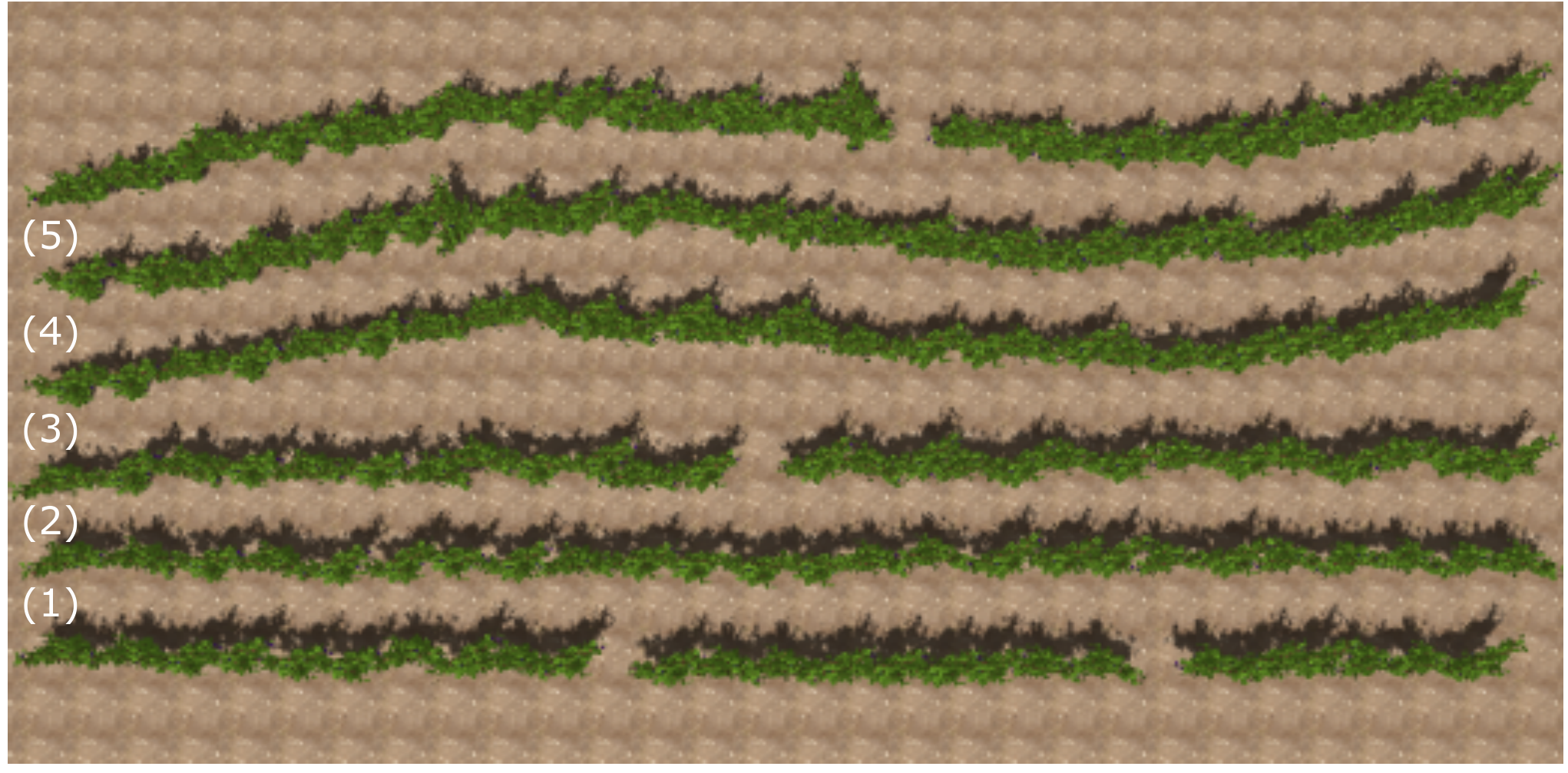}
        \label{fig:test_env}
    }
    \caption{Training (\textbf{a}) and testing (\textbf{b}) simulation environments. Intentional gaps and curves are introduced in (\textbf{b}) to provide a challenging environment for the agent. Testing rows are numbered as referenced in the results section.}
    \label{fig:test_train_env}
\end{figure}

We design two different custom environments, one for training and one for testing, completely from scratch, with the aim of reproducing the geometry of real vineyards (Fig. \ref{fig:test_train_env}). The inter-row distance ranges from 1.5m to 2.0m, while plants in the same row have a distance between them that ranges from 0.7m to 1.0m.  We further increase the challenge in the testing environment organizing the plants along both straight and curved lines and creating gaps in the rows, as shown in Fig. \ref{fig:test_env}. Plant gaps are a common condition in real-world vineyards and, as reported in previous studies \cite{aghi2021deep}, they represent the most common failure situation for visual-based guidance algorithms, since they may easily fool an agent to leave the main row path.

\begin{figure*}[t]
\centering
\includegraphics[width=0.9\textwidth]{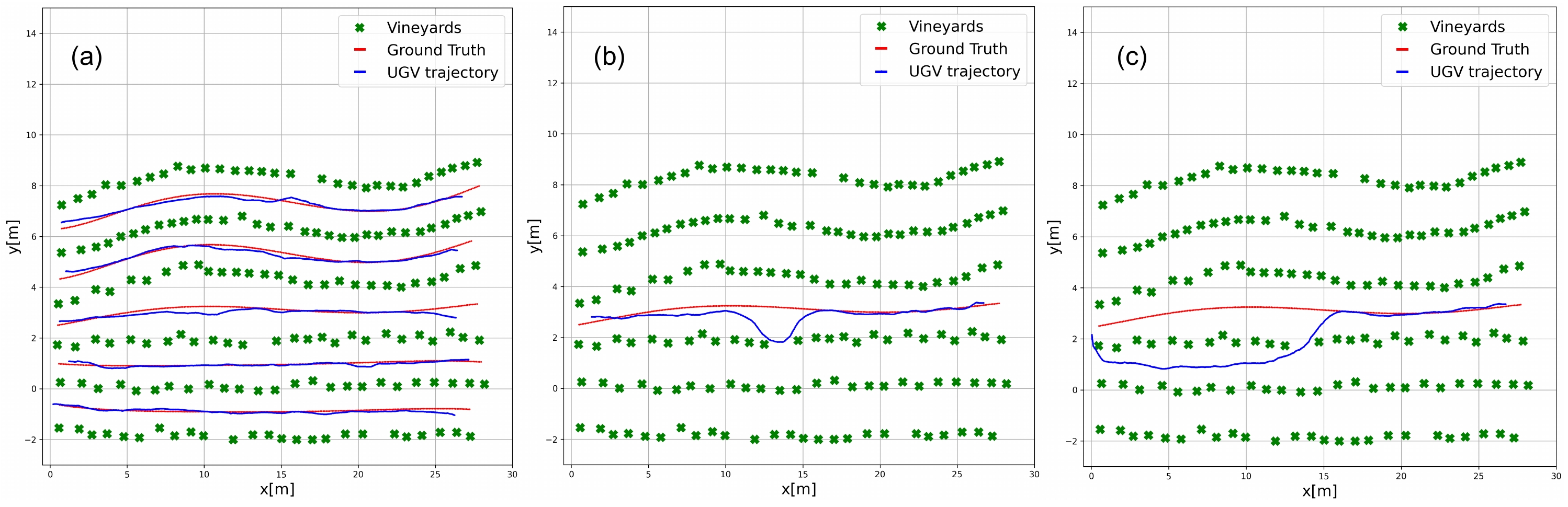}
\caption{Results achieved by the agent in the testing environment. (\textbf{a}) resumes the performance on all the rows, where the agent follows a trajectory very close to the optimal one. In (\textbf{b}), the agent recovers from a critical situation given by gaps in the row. In (\textbf{c}), the agent fails and switches to the adjacent row but still reaches the end of the vineyard without collisions.}
\label{fig:tests}
\end{figure*}

\section{EXPERIMENTS AND RESULTS}
\label{sec:results_sec}
In this section we present the experimental session followed to validate the proposed methodology, with the aim of answering to the following questions:

1) Is the agent able to successfully accomplish the task of vineyard navigation in an unseen environment?

2) How well can the agent guide the robot through the vineyard? In particular, our main objective is to evaluate the quality of the produced trajectory in terms of number of collisions, centrality, and average velocity commands.

3) How well can the agent generalize to new testing conditions such as a different robotic platform or an increased level of noise in the depth image?

To this end, we firstly report the performance of our navigation agent in the testing environment evaluating the quality of the trajectory and the overall behaviour of the robot on several repeated tests. Moreover, we test the generalization and robustness properties of our trained policy, varying the robotic platform and increasing the noise in the depth images. Finally, we provide an analysis of the real-time performance on different computational systems.

\subsection{Vineyard Navigation Test}
\label{subsec:test_results}
We test our navigation methodology on the five vineyard rows of the testing environment of Fig. \ref{fig:test_env}: two straight rows, two curved rows and a hybrid row. We perform a total amount of ten tests for each row, half in the forward direction (F) and half in the reverse direction (R). We repeat tests multiple times to better validate our approach and obtain more consistent results.

The main approach to validate autonomous navigation algorithms is to compare the trajectory followed by the platform with a ground truth path. We compute the median line for each vineyard row in the testing environment, performing a mean operation between two adjacent rows, then we fit the obtained median points with a fifth-order polynomial in order to achieve an accurate ground truth line. Afterwards, we compare the robot trajectory with the computed ground truth lines in order to estimate its Mean Absolute Error (MAE) and Root Mean Square Error (RMSE). Moreover, for each test we report the number of trials in which the UGV is able to successfully reach the end of the row.

Table \ref{tab:test_results} presents the complete quantitative evaluation of the experiment. Despite the difficulty of the test environment, the learned policy demonstrates the ability of correctly guiding the robot through different rows. As expected, the agent performs slightly better on the straight test rows (1 and 2) compared to the curved ones (4 and 5), since the conditions are closer to the training scenario. That is proved by the lower MAE and RMSE values. Failures mostly occur in the hybrid test row 3R (2 successes out of 5) and in the curved row 5F (3 successes out of 5), due to the presence of a wide gap between the plants. However, the agent is able to correctly cope with gaps in the other tests. Fig. \ref{fig:tests} presents some examples of trajectories obtained during the test phase, particularly relevant to the gap handling. During the tests shown in Fig. \ref{fig:tests}a, the agent is able to follow a path very close to the optimal one. In Fig. \ref{fig:tests}b, the gap causes a suboptimal behavior, but the agent is still able to recover the correct path. On the other hand, in  Fig. \ref{fig:tests}c, an example of a failure is presented. It is worth citing that, despite following the wrong path, the agent is still able to avoid collisions. The overall results report an MAE of 0.114m, an RMSE of 0.136m and a success rate of 45/50 total runs, which is a considerably positive result.

\begin{table}[t]
\centering
\caption{Comparison of agent generalization capabilities on a different robotic platform. We test the rovers on both straight (1F) and curved (4F) rows, reporting the average results among 3 runs.}
\label{tab:rover_test}
\resizebox{0.5\columnwidth}{!}{%
\begin{tabular}{@{}cccccccc@{}}
\toprule
Row & Rover & Success & $T_{avg}$ {[}s{]} & MAE {[}m{]} & RMSE {[}m{]} \\ \midrule
\multirow{2}{*}{Straight} & Jackal & 3/3 & 78 & 0.079 & 0.088 \\
                          & Husky  & 3/3 & 69 & 0.081 & 0.098 \\ \midrule
\multirow{2}{*}{Curved}    & Jackal & 3/3 & 78 & 0.085 & 0.101 \\
                          & Husky  & 2/3 & 88 & 0.112 & 0.138 \\ \bottomrule
\end{tabular}%
}
\end{table}

\subsection{Generalization, Robustness and Real-Time Performance}
\label{subsec:robustness_sec}

\begin{table}[t]
\centering
\caption{Success rate of the agent tested with increasing noise factor in both straight (1F) and curved (4F) vineyard rows with three trials for each test run.}
\label{tab:robustness}
\resizebox{0.5\columnwidth}{!}
{
\begin{tabular}{@{}lccccc@{}}
\toprule
\begin{tabular}[c]{@{}l@{}}Noise Factor\end{tabular} & 2 & 4 & 6 & 8 & 10 \\ \midrule
\begin{tabular}[c]{@{}l@{}}Success (Straight Row)\end{tabular} & 3/3 & 3/3 & 3/3 & 2/3 & 0/3 \\ \midrule
\begin{tabular}[c]{@{}l@{}}Success (Curved Row)\end{tabular} & 3/3 & 3/3 & 3/3 & 3/3 & 0/3 \\ \bottomrule
\end{tabular}%
}
\end{table}

\begin{figure}[ht]
\centering
\includegraphics[width=0.8\columnwidth]{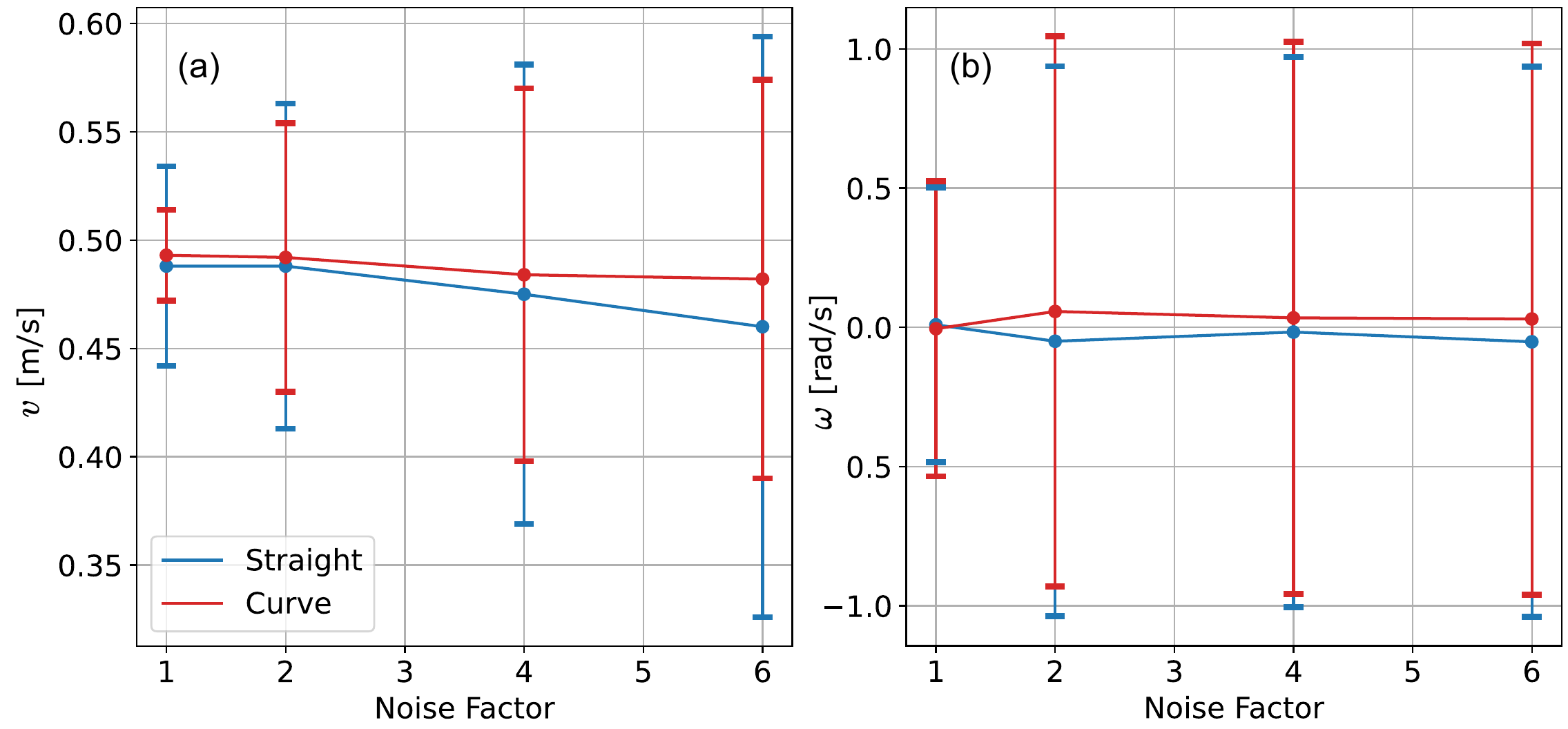}
\caption{Model robustness to noise. We report the mean and standard deviation for $v$ \textbf{(a)} and $\omega$ \textbf{(b)}, highlighting how the predictions of the policy network tend to oscillate more when the noise level increases. However, the agent is able to reach the end of the row in almost all the cases even with high levels of noise.}
\label{fig:noise}
\end{figure}

We conduct further experiments to evaluate the generalization and robustness properties of the trained policy varying two separate impacting conditions: the rover platform and the amount of noise in the depth image. These additional tests are performed on two different rows of the testing environments: one straight (Row 1F) and one curved (Row 4F). We repeat each test three times for a consistent comparison in both the configurations. Moreover, we analyze the inference timing of the proposed agent with different computational systems in order to report the ability of the actual platforms to run the policy on-board in real-time.

\begin{description}[leftmargin=0pt]
\item[Different Robotic Platform]
We choose the Husky UGV\footnote{https://clearpathrobotics.com/husky-unmanned-ground-vehicle-robot/} to test our policy with a different platform. The Husky URDF model is also provided by Clearpath Robotics. Husky has an overall size Length $\times$ Width $\times$ Height equal to $990 \times 670 \times 390$ mm, compared to the smaller Jackal size of $508 \times 430 \times 250$ mm. The larger footprint and the increased height of the camera constitute sensible variations for the agent, which needs to select actions accordingly to manage collision avoidance. Overall, the results presented in Table \ref{tab:rover_test} can be considered successful: the agent is able to guide the new UGV to the end of the vineyard row with a success rate of $3/3$ in the straight case and $2/3$ in the curved case. Despite the slight degradation in the MAE and RMSE of the trajectory and a longer time required in the curved row, the agent demonstrates to generalize well with different platforms, taking into account the considerable differences between the two robotic platforms.

\item[Depth Image Noise]
Real-world depth images often present unpredictable noisy behaviours. Aware of that, we investigate the ability of our solution to resist to heavier noise in the observations, by gradually applying a multiplicative factor to the noise in the images. The trained policy demonstrates to be robust to noise by successfully reaching the end of the row until using a maximum noise factor of $10$. We consider noise multiplication factors of $2, 4, 6, 8, 10$, obtaining the success rate presented in Table \ref{tab:robustness}. Moreover, we show the effect of the increased noise in the velocity commands generated by the agent in Fig. \ref{fig:noise}. The mean value of the linear velocity commands is reduced proportionally to the increase of the perturbation in the depth image, demonstrating how the agent prioritizes a safe traveling with respect to speed. On the other hand, the standard deviation in both velocities result to be much higher, suggesting uncertainty and oscillation in the robot behaviour. Nonetheless, considering the degradation level used in the depth image, the obtained performances are strongly encouraging in terms of noise tolerance and robustness.

\begin{table}[t]
\centering
\caption{Real-time inference performance of the actor policy network on different hardware configurations. Each test is reported with a statistic on 100 independent trials.}
\label{tab:realtime}
\resizebox{0.5\columnwidth}{!}{%
\begin{tabular}{@{}cccc@{}}
\toprule
Platform & CPU  & Model & T {[}ms{]} \\ \midrule
\multirow{2}{*}{Jackal} & \multirow{2}{*}{Intel i3-4330TE@2.40GHz}
                        & TF & $6.20\pm1.50$ \\
                        & & TFLite & $1.96\pm0.11$ \\\midrule
\multirow{2}{*}{Husky}  & \multirow{2}{*}{Intel i7-6700TE@2.40GHz}
                        & TF & $4.42\pm1.35$ \\
                        & & TFLite & $1.24\pm0.21$ \\\midrule
\multirow{2}{*}{NUC}    & \multirow{2}{*}{Intel i5-1145G7@2.60GHz}
                        & TF & $2.80\pm0.62$ \\
                        & & TFLite & $0.63\pm0.39$ \\\bottomrule
\end{tabular}%
}
\end{table}

\item[Real-Time Performance]
As additional analysis, we test inference timings of the actor policy network on three different computing hardware to investigate its performance on real platforms. In particular, we test the computers mounted on Jackal and Husky UGVs, together with an Intel NUC mini PC, an emerging solution as on-board computing hardware for robotics platforms. For our test, we apply network optimizations with the TensorFlow Lite\footnote{https://www.tensorflow.org/lite} library, obtaining the float32 \textit{.tflite} converted model. Results presented in Tab. \ref{tab:realtime} show how all the considered hardware configurations can easily reach real-time performance and that TFLite conversion heavily decreases inference timings.

\end{description}


\section{Conclusions}
In this work, we proposed a learning-based methodology to tackle the problem of position-agnostic autonomous navigation in vineyard rows. We demonstrated that Deep Reinforcement Learning represents a competitive approach to learn a navigation policy for a mobile robot in complex constrained environments. We conceived our solution to handle the absence of a precise positioning information, which is often unavailable in vineyards. With our extensive experimentation we showed that our trained policy can safely guide the robot through unseen straight and curved vineyard rows, generating considerably stable central trajectory. Moreover, we investigated the generalization and robustness properties of our solution by testing the agent with a different UGV model and heavy depth image perturbations. Our findings pave the way for further studies on learning-based solutions for autonomous navigation in precision agriculture applications. Future works will involve a more precise modeling of depth camera noise and real-world testing of the algorithm, also evolving the agent to handle inter-rows navigation and integrating it with a global waypoints planner.

\section*{Acknowledgments}
This work has been developed with the contribution of the Politecnico di Torino Interdepartmental Centre for Service Robotics (PIC4SeR) and SmartData@Polito.

\bibliographystyle{unsrt}  
\bibliography{mybib}

\end{document}